\pdfoutput=1

\documentclass[11pt]{article}

\usepackage[preprint]{acl}

\usepackage{times}
\usepackage{latexsym}

\usepackage[T1]{fontenc}

\usepackage[utf8]{inputenc}

\usepackage{microtype}

\usepackage{inconsolata}

\usepackage{graphicx}

\usepackage{natbib}

\usepackage{amsmath}
\usepackage{amssymb}

\usepackage[ruled,vlined]{algorithm2e}

\usepackage{arydshln}

\usepackage{enumitem}

\usepackage{xcolor}
\usepackage{soul} 

\definecolor{myRed}{HTML}{FFB2B3}
\definecolor{myGreen}{HTML}{B3DAB1}

\title{CIKT: A Collaborative and Iterative Knowledge Tracing Framework with Large Language Models}

\author{
  Runze Li, Siyu Wu, Jun Wang, Wei Zhang\thanks{Corresponding author}  \\
  School of Computer Science and Technology, East China Normal University \\
  \texttt{51265901021@example.edu.cn, 52265901039@stu.ecnu.edu.cn}\\
  \texttt{wongjun@gmail.com, zhangwei.thu2011@gmail.com}
}

\begin{document}
\maketitle

\begin{abstract}
Knowledge Tracing (KT) aims to model a student’s learning state over time and predict their future performance. 
However, traditional KT methods often face challenges in explainability, scalability, and effective modeling of complex knowledge dependencies. While Large Language Models (LLMs) present new avenues for KT, their direct application often struggles with generating structured, explainable student representations and lacks mechanisms for continuous, task-specific refinement. To address these gaps, we propose Collaborative Iterative Knowledge Tracing (CIKT), a framework that harnesses LLMs to enhance both prediction accuracy and explainability. CIKT employs a dual-component architecture: an Analyst generates dynamic, explainable user profiles from student historical responses, and a Predictor utilizes these profiles to forecast future performance. 
The core of CIKT is a synergistic optimization loop.
In this loop, the Analyst is iteratively refined based on the predictive accuracy of the Predictor, which conditions on the generated profiles, and the Predictor is subsequently retrained using these enhanced profiles.
Evaluated on multiple educational datasets, CIKT demonstrates significant improvements in prediction accuracy, offers enhanced explainability through its dynamically updated user profiles, and exhibits improved scalability.
Our work presents a robust and explainable solution for advancing knowledge tracing systems, effectively bridging the gap between predictive performance and model transparency.
\end{abstract}

\section{Introduction}
\label{sec:introduction}

Knowledge Tracing (KT)~\cite{corbett1994knowledge} is a foundational task in educational data mining and intelligent tutoring systems, aiming to model a student's evolving knowledge state from their historical learning interactions to accurately predict future performance, thereby facilitating personalized learning and targeted interventions. While early approaches like Bayesian Knowledge Tracing (BKT)~\cite{corbett1994knowledge} and its extensions~\cite{pardos2011kt, pardos2010modeling} offered interpretable parameters, they often struggled with the complex temporal dependencies of learning processes. Deep learning-based KT (DLKT) models subsequently emerged, significantly advancing predictive accuracy. Pioneering models such as Deep Knowledge Tracing (DKT)~\cite{piech2015deep} with Recurrent Neural Networks, and memory-augmented architectures like DKVMN~\cite{zhang2017dynamic}, laid crucial groundwork~\cite{liu2019ekt, nagatani2019augmenting}. Further advancements, including Transformer-based models like SAKT~\cite{pandey2019self}, AKT~\cite{ghosh2020context}, and LPKT~\cite{shen2021learning} (which incorporated cognitive dynamics), alongside innovations like FoLiBi's linear forgetting mechanisms~\cite{im2023forgetting} and the integration of side information~\cite{wang2021temporal, pandey2020rkt} or graph structures~\cite{nakagawa2019graph, yang2025mahkt, wang2025multi}, have continued to push KT performance boundaries.

Despite these significant strides in predictive power, achieving robust explainability remains a persistent challenge in the DLKT landscape~\cite{minn2022interpretable, zhao2020interpretable}. Although various strategies have been explored—from inherently interpretable components~\cite{zhang2017dynamic, shen2021learning} and post-hoc analyses~\cite{scruggs2019extending} to aligning models with learning theories~\cite{chen2023improving, cui2024interpretable}—many DLKT models remain substantially opaque. This lack of transparency can hinder their adoption and trustworthiness in high-stakes educational settings where understanding the model's reasoning is crucial.

The transformative capabilities of LLMs, demonstrated across specialized domains like scientific discovery~\cite{pyzer2022accelerating, merchant2023scaling} and automated research assistance~\cite{wang2024autosurvey, wang2024scimon, lu2024ai, huang2023mlagentbench, tyser2024ai}, offer promising new avenues for addressing KT's dual challenges. However, directly applying general-purpose LLMs to the nuanced task of knowledge tracing introduces distinct difficulties: (1) eliciting structured, interpretable representations of dynamic student knowledge states beyond mere task-specific predictions; (2) optimizing LLM behavior for KT without abundant, explicit preference signals or fine-grained supervision for explainability; and (3) resolving the inherent tension between maximizing predictive accuracy and maintaining KT process explainability. Moreover, many current LLM applications operate statically post-deployment, lacking mechanisms for continuous self-improvement based on domain-specific feedback.

To address these multifaceted limitations, we propose Collaborative Iterative Knowledge Tracing (CIKT), a framework architected around two core LLM-based components: an Analyst that generates structured, interpretable student profiles from historical responses, and a Predictor that leverages these profiles for future performance forecasting. The cornerstone of CIKT is an iterative learning strategy employing Kahneman-Tversky Optimization (KTO)~\cite{ethayarajh2024kto}. This mechanism facilitates reciprocal enhancement: the Predictor's accuracy, conditioned on Analyst-generated profiles, provides reinforcement-style feedback to progressively refine the Analyst. Subsequently, the Predictor is retrained with these enhanced profiles, completing a collaborative optimization loop. Both the Analyst and Predictor are built upon a large-scale pre-trained language model backbone, ensuring flexibility and powerful representation learning.

The major contributions of this paper are summarized as follows:
\begin{itemize}[leftmargin=*, topsep=0pt, partopsep=0pt, itemsep=0pt, parsep=0pt] 
    \item We propose a \textbf{collaborative knowledge tracing framework} that explicitly models student knowledge states via an Analyst and utilizes these dynamic profiles for predictive tasks through a Predictor.
    \item We introduce \textbf{an iterative optimization strategy} based on reinforcement-style feedback, enabling mutual refinement between the Analyst and Predictor to improve both the quality of generated profiles and overall predictive performance.
    \item We conduct extensive experiments on multiple educational datasets, demonstrating that our CIKT framework outperforms existing KT models in predictive accuracy while simultaneously offering enhanced explainability through its generated student profiles.
\end{itemize}

\section{Related Work}
Knowledge tracing (KT) \cite{corbett1994knowledge}, a key task in educational data mining, models students’ evolving knowledge states to predict future performance. 
Early models like Bayesian Knowledge Tracing (BKT) \cite{corbett1994knowledge} used binary mastery variables and explainable parameters for learning/forgetting dynamics. 
Extensions to BKT \cite{pardos2011kt,pardos2010modeling} and other machine learning methods \cite{pavlik2009performance} aimed to improve accuracy and flexibility. 
However, these models struggled with complex temporal dependencies and latent interactions.

Deep learning-based knowledge tracing (DLKT) models emerged to overcome these limitations. 
Deep Knowledge Tracing (DKT) \cite{piech2015deep} notably used RNNs to learn latent representations from student interactions. 
DKVMN \cite{zhang2017dynamic} later enhanced the structure by using key-value memory networks for concept mastery tracking. 
These pioneering DLKT models established a foundation for later work \cite{liu2019ekt,nagatani2019augmenting,shen2021learning}.

Recent advances leverage attention mechanisms and Transformers to better model long-range dependencies. 
Models like SAKT \cite{pandey2019self}, AKT \cite{ghosh2020context}, LPKT \cite{shen2021learning} (with memory-aligned gates), and FoLiBi \cite{im2023forgetting} (with linear forgetting) improved accuracy and explainability by modeling contextual and cognitive dynamics. 
Integrating side information like temporal or contextual features \cite{wang2021temporal,pandey2020rkt} also enhanced KT performance. 
Graph-based methods \cite{nakagawa2019graph,yang2025mahkt,wang2025multi} model concept and interaction dependencies for better knowledge representation.

Explainability remains a key DLKT concern despite these developments. Efforts include inherently explainable architectures \cite{zhang2017dynamic,shen2021learning,minn2022interpretable}, post-hoc analysis of trained models (e.g., attention weights) \cite{zhao2020interpretable,scruggs2019extending}, and integrated modules like attention or cognitive mechanisms. 
However, these methods often face generalizability issues and task-specific design dependencies.
Other works target explainability by aligning models with learning theories \cite{chen2023improving,cui2024interpretable}.
Still, most DLKT models remain opaque, posing audit challenges in high-stakes education.

The adaptation of Large Language Models (LLMs) for specialized applications in various vertical domains shows considerable promise beyond general-purpose tasks. For instance, their capabilities are harnessed for nuanced information processing, such as automating scientific literature retrieval~\cite{wang2024autosurvey}, generating domain-specific survey papers~\cite{wang2024scimon}, aiding complex data analysis in scientific discovery (e.g., material discovery~\cite{pyzer2022accelerating, merchant2023scaling}), supporting prompt-driven research pipelines~\cite{lu2024ai}, evaluating specialized content like scientific papers~\cite{tyser2024ai}, and assisting in domain-specific coding solutions~\cite{huang2023mlagentbench}. While these domain-specific adaptations often achieve notable performance through fine-tuning or sophisticated prompting, a common limitation is their static operation post-deployment; they typically lack embedded processes for continuous self-iteration and performance enhancement based on ongoing, domain-specific feedback. Addressing this crucial gap, our work proposes a novel collaborative iterative optimization framework specifically designed to empower LLMs to continuously refine their own effectiveness for the specialized task at hand.

\section{Methodology}
\label{sec:methodology}

\begin{figure*}[htbp]
  \centering
  \includegraphics[width=0.8\textwidth]{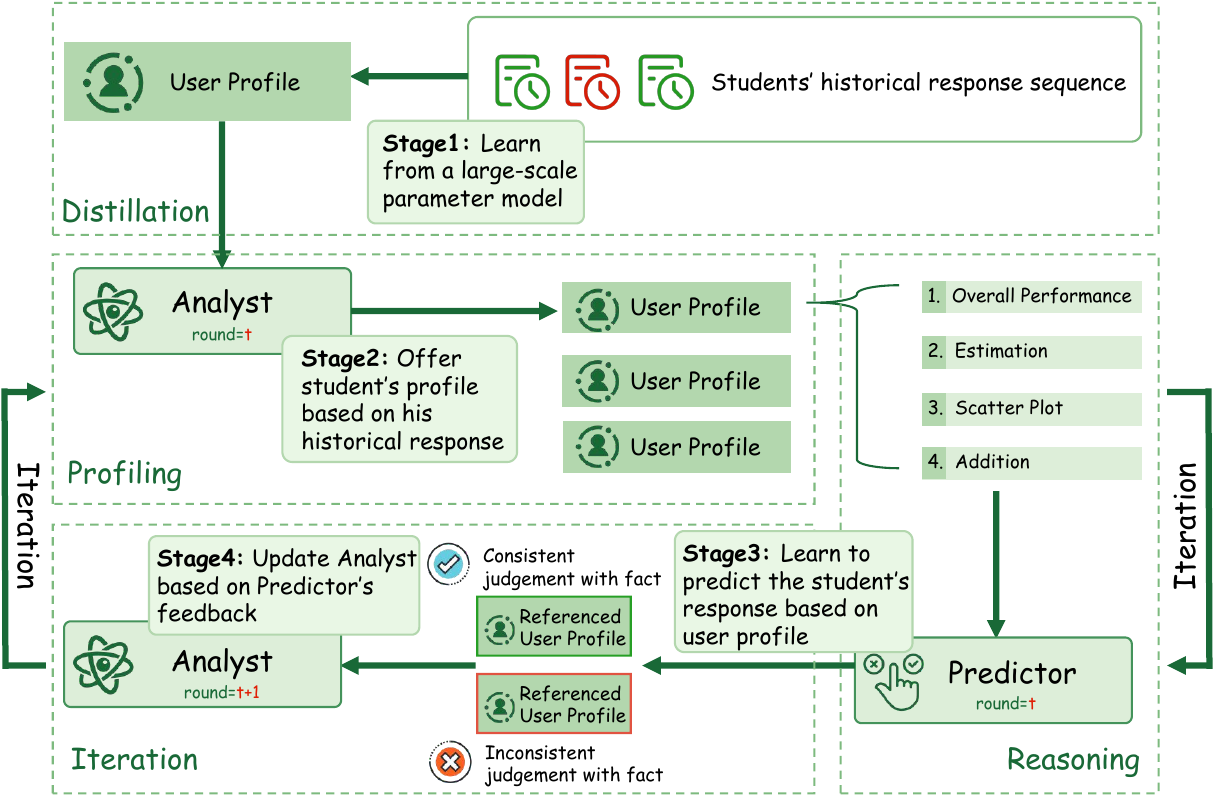}
  \caption{\textbf{The CIKT framework}, illustrating the collaborative four-stage process involving the Analyst and Predictor: distillation, profiling, reasoning, and iteration.}
  \label{fig:training_pipeline}
\end{figure*}

\subsection{Overall Framework}
To enhance both the accuracy and explainability of KT, we propose CIKT, a framework leveraging the capabilities of LLMs. CIKT revolves around two core intelligent components: an Analyst responsible for generating rich, structured student profiles from historical interaction data, and a Predictor that utilizes these profiles, alongside interaction history, to forecast student performance. The synergy between the Analyst and Predictor is cultivated through a meticulously designed four-stage iterative process, illustrated in Figure~\ref{fig:training_pipeline}. This cycle begins with \textbf{Distillation}, where the Analyst learns foundational profiling capabilities from curated annotations initially provided by a large teacher model. Next, in the \textbf{Profiling} stage, the trained Analyst generates comprehensive user profiles for student data. Subsequently, during \textbf{Reasoning}, the Predictor is trained to predict outcomes using these profiles and historical interactions. Finally, the \textbf{Iteration} stage employs a refinement loop where feedback from the Predictor's performance, guided by Kahneman-Tversky Optimization (KTO)~\cite{ethayarajh2024kto} principles, is used to further optimize the Analyst via reinforcement learning. This iterative process facilitates mutual improvement, enhancing both the quality of the generated profiles and the accuracy of predictions.

\subsection{Stage 1: Distillation}

This initial Distillation stage aims to endow the Analyst with the foundational capability to generate structured and informative user profiles from raw student interaction data. 
The process begins by leveraging a large-scale teacher model, $\text{LLM}_{\text{teacher}}$ (e.g., GPT-4o~\cite{hurst2024gpt}), to process historical interaction sequences from a subset of students. For a student $s$, their sequence is denoted as $\mathcal{S}_s = \{(e_1, r_1), (e_2, r_2), \dots, (e_N, r_N)\}$, where $e_i$ is the $i$-th exercise and $r_i \in \{0, 1\}$ its binary correctness. The $\text{LLM}_{\text{teacher}}$ produces initial textual profiles:
\begin{equation}
    \mathbf{p}_{s, \text{teacher}} = \text{LLM}_{\text{teacher}}(\mathcal{S}_s)
    \label{eq:raw_teacher_profile}
\end{equation}
These profiles, $\mathbf{p}_{s, \text{teacher}}$, are designed as textual outputs capturing the student's knowledge state, including aspects like mastery levels across knowledge concepts, inferred learning patterns, and potential difficulties. Subsequently, these profiles undergo a manual curation process where their format and content are reviewed and corrected; only selected, high-quality profiles, denoted as $\mathbf{p}^*_{s, \text{teacher}}$, that accurately reflect student understanding are retained for training.

The Analyst, parameterized by $\theta_A$ and based on our chosen backbone LLM architecture, is then fine-tuned via supervised learning using pairs of student sequences ($\mathcal{S}_s$) and their corresponding curated teacher profiles ($\mathbf{p}^*_{s, \text{teacher}}$) from a training set $\mathcal{D}_{\text{train}}$. The Analyst learns to map an input sequence $\mathcal{S}_s$ to its own profile generation $\mathbf{p}_{s, \text{analyst}}$:
\begin{equation}
    \mathbf{p}_{s, \text{analyst}} = \text{Analyst}(\mathcal{S}_s; \theta_A)
    \label{eq:analyst_profile_distill}
\end{equation}
The training objective is to minimize a distillation loss, $\mathcal{L}_{\text{Distill}}$. Given that the profiles are textual, this loss is formulated as a token-level cross-entropy ($\mathcal{L}_{\text{CE}}$) between the Analyst-generated profiles and the curated teacher profiles:
\begin{equation}
    \mathcal{L}_{\text{Distill}}(\theta_A) = \sum_{s \in \mathcal{D}_{\text{train}}} \mathcal{L}_{\text{CE}}(\mathbf{p}^*_{s, \text{teacher}}, \mathbf{p}_{s, \text{analyst}})
    \label{eq:distill_loss_ce}
\end{equation}
This supervised distillation phase equips the Analyst with a robust initial model for generating meaningful user profiles.

\begin{table*}[htbp]
    \centering
    \resizebox{0.9\textwidth}{!}{
    \begin{tabular}{l c c c c c c c c c c}
        \hline
        \textbf{Dataset} & \multicolumn{3}{c}{\textbf{ASSIST2009}} & \multicolumn{3}{c}{\textbf{ASSIST2012}} & \multicolumn{3}{c}{\textbf{Eedi}} \\ 
        Metrics & ACC & $\text{ACC}_{len>15}$ & F1 & ACC & $\text{ACC}_{len>15}$ & F1 & ACC & $\text{ACC}_{len>15}$ & F1 \\
        \hline
        DKT & 0.737 & 0.744 & 0.808 & 0.745 & 0.737 & 0.828 & 0.712 & 0.707 & 0.814 \\
        AKT & 0.725 & 0.716 & 0.801 & 0.744 & 0.725 & 0.821 & 0.719 & 0.702 & 0.807 \\
        SAKT & 0.743 & 0.755 & 0.810 & 0.743 & 0.747 & 0.830 & 0.722 & 0.725 & 0.816 \\
        LPKT & 0.738 & 0.750 & 0.812 & 0.731 & 0.751 & \underline{0.832} & 0.728 & \underline{0.729} & 0.818 \\
        IKT & 0.726 & 0.710 & 0.802 & 0.745 & 0.741 & 0.807 & 0.719 & 0.701 & 0.793 \\
        DIMKT & \underline{0.749} & 0.748 & \underline{0.810} & \underline{0.747} & \underline{0.756} & 0.816 & \underline{0.733} & 0.727 & \underline{0.819} \\
        DKVMN & 0.724 & 0.729 & 0.793 & 0.735 & 0.727 & 0.760 & 0.719 & 0.715 & 0.766 \\
        \hline
        GPT-4o & 0.732 & \underline{0.756} & 0.794 & 0.720 & 0.619 & 0.802 & - & - & - \\
        Deepseek-R1 & 0.669 & 0.665 & 0.762 & 0.646 & 0.671 & 0.770 & - & - & - \\
        \hline
        CIKT-Llama3.1-8B & 0.775 & \textbf{0.781} & \textbf{0.827} & 0.774 & 0.784 & 0.847 & 0.770 & 0.775 & 0.834 \\
        CIKT-Qwen2.5-7B & \textbf{0.777} & 0.778 & 0.820 & \textbf{0.780} & \textbf{0.790} & \textbf{0.852} & \textbf{0.777} & \textbf{0.781} & \textbf{0.836} \\
        \hline
        improv. & +3.74\% & +4.13\% & +2.10\% & +4.42\% & +4.50\% & +2.40\% & +6.00\% & +7.13\% & +2.08\% \\
        \hline
    \end{tabular}}
    \caption{Results of the main experiments.}
    \label{tab:main}
\end{table*}

\subsection{Stage 2: Profiling}
\label{sec:profiling}

Following the initial foundation building in the Distillation stage, the Analyst, now equipped with its learned parameters $\theta_A$, is applied to generate user profiles for all relevant student data. The primary objective of this Profiling stage is to transform raw student historical interaction sequences into rich, structured profile representations that will inform the subsequent prediction tasks.

For each student $s$ with a historical interaction sequence $\mathcal{S}_s = \{(e_1, r_1), (e_2, r_2), \dots, (e_N, r_N)\}$, the trained Analyst synthesizes a corresponding user profile $\mathbf{p}_s$. This process can be represented as:
\begin{equation}
    \mathbf{p}_s = \text{Analyst}(\mathcal{S}_s; \theta_A)
    \label{eq:analyst_profiling_application}
\end{equation}
where $\mathbf{p}_s$ is the textual profile generated by the Analyst based on the student's historical interactions. This profiling step is systematically applied across the entire dataset, including the training, validation, and test sets. The resulting set of user profiles $\{\mathbf{p}_s\}$ serves as a crucial augmented input, alongside the original interaction sequences $\{\mathcal{S}_s\}$, for training and evaluating the Predictor in the subsequent Reasoning stage (Section~\ref{sec:reasoning}). The quality and informativeness of these profiles are paramount for the Predictor's ability to make accurate and nuanced performance forecasts.

\subsection{Stage 3: Reasoning}
\label{sec:reasoning}
The "Reasoning" stage centers on training the Predictor, parameterized by $\theta_P$. Its objective is to accurately forecast a student's performance $y_{s,t}$ (where $y_{s,t} \in \{0,1\}$ indicates binary correctness) on a subsequent learning exercise $e_t$. To achieve this, the Predictor utilizes a combination of the student's historical interaction sequence $\mathcal{H}_{s,t-1} = \{(e_1, r_1), \dots, (e_{t-1}, r_{t-1})\}$ (where $e_i$ is an exercise and $r_i$ its correctness), the corresponding user profile $\mathbf{p}_{s,t-1}$ generated by the Analyst (i.e., $\mathbf{p}_{s,t-1} = \text{Analyst}(\mathcal{H}_{s,t-1}; \theta_A)$), and information pertaining to the target exercise $e_t$. The Predictor then outputs the predicted probability of a correct response:
\begin{equation}
    \hat{y}_{s,t} = \text{Predictor}(\mathcal{H}_{s,t-1}, \mathbf{p}_{s,t-1}, e_t; \theta_P)
    \label{eq:predictor_output}
\end{equation}

Training of the Predictor is conducted via supervised fine-tuning. Given a training set $\mathcal{D}_{\text{train}}$ comprising instances of $(\mathcal{H}_{s,t-1}, \mathbf{p}_{s,t-1}, e_t, y_{s,t})$, the parameters $\theta_P$ are optimized by minimizing the binary cross-entropy loss function, $\mathcal{L}_{\text{Predict}}$. This loss encourages the predicted probabilities $\hat{y}_{s,t}$ to closely align with the true outcomes $y_{s,t}$:
\begin{equation}
\label{eq:predictor_loss}
\begin{split}
    \mathcal{L}_{\text{Predict}}(\theta_P) = & -\sum_{(s,t) \in \mathcal{D}_{\text{train}}} \Biggl[ y_{s,t} \log(\hat{y}_{s,t}) \\
    & \hspace{2em} + (1 - y_{s,t}) \log(1 - \hat{y}_{s,t}) \Biggr]
\end{split}
\end{equation}
This process enables the Predictor to learn complex relationships between past learning activities, the summarized knowledge state encapsulated in the profile, and future performance, thereby effectively reasoning to arrive at its predictions.

\begin{table}[htbp]
\centering
\small
\begin{tabular}{lcccc}
\hline
\textbf{Dataset} & ASSIST2009 & ASSIST2012 & Eedi \\
\hline
\textbf{\# Responses} & 0.4m & 2.7m & 17.8m  \\
\textbf{\# Sequences} & 8.3k & 67.1k &  475.4k \\
\textbf{\# Questions} & 6.9k & 53.1k & 2.7k \\
\textbf{\# Concepts} & 200 & 265 & 386 \\
\hline
\end{tabular}
\caption{Statistics of the preprocessed datasets.}
\label{tab:dataset_statistics}
\end{table}

\begin{table*}[htbp]
    \centering
    \resizebox{0.9\textwidth}{!}{
    \begin{tabular}{l c c c c c c c c c c}
        \hline
        \textbf{Dataset} & \multicolumn{3}{c}{\textbf{ASSIST2009}} & \multicolumn{3}{c}{\textbf{ASSIST2012}} & \multicolumn{3}{c}{\textbf{Eedi}} \\ 
        Metrics & ACC & $\text{ACC}_{len>15}$ & F1 & ACC & $\text{ACC}_{len>15}$ & F1 & ACC & $\text{ACC}_{len>15}$ & F1 \\
        \hline
        CIKT-Llama3.1-8B & \textbf{0.775} & \textbf{0.781} & \textbf{0.827} & \textbf{0.774} & \textbf{0.784} & \textbf{0.847} & \textbf{0.770} & \textbf{0.775} & \textbf{0.834} \\
        \hline
        \textit{train} &  &  &  &  &  &  &  &  &  \\
        w/o Iteration \& Cooperation & 0.766 & 0.772 & 0.818 & 0.767 & 0.770 & 0.841 & 0.768 & 0.747 & 0.830 \\
        w/o Iteration & 0.772 & 0.777 & 0.824 & 0.756 & 0.755 & 0.833 & 0.755 & 0.767 & 0.830  \\
        \hdashline
        \textit{inference} &  &  &  &  &  &  &  &  &  \\
        w/o Profile & 0.755 & 0.750 & 0.805 & 0.760 & 0.765 & 0.835 & 0.760 & 0.700 & 0.830 \\
        \hline
        \hline
        CIKT-Qwen2.5-7B & \textbf{0.777} & \textbf{0.778} & \textbf{0.820} & \textbf{0.780} & \textbf{0.790} & \textbf{0.852} & \textbf{0.777} & \textbf{0.781} & \textbf{0.836} \\
        \hline
        \textit{train} &  &  &  &  &  &  &  &  &  \\
        w/o Iteration \& Cooperation & 0.765 & 0.770 &0.812 & 0.766 & 0.777 & 0.843 & 0.761 & 0.756 & 0.826 \\
        w/o Iteration & 0.768 & 0.772 & 0.814 & 0.775 & 0.747 & 0.851 & 0.775 & 0.771 & 0.833 \\
        \hdashline
        \textit{inference} &  &  &  &  &  &  &  &  &  \\
        w/o Profile & 0.749 & 0.750 & 0.798 & 0.768 & 0.732 & 0.844 & 0.765 & 0.757 & 0.835 \\
        \hline
    \end{tabular}}
    \caption{Results of the ablation experiments. "CIKT w/o Iteration" removes the iterative refinement but keeps the cooperative structure trained in a single pass. "CIKT w/o Iteration \& Cooperation" removes both iteration and the cooperative structure, representing a more direct LLM fine-tuning. "CIKT w/o Profile (Inference)" means the full model was trained, but profiles were withheld from the Predictor during inference.}
    \label{tab:ablation}
\end{table*}

\subsection{Stage 4: Iteration}
\label{sec:iteration_kto}

The Iteration stage is pivotal to our CIKT framework's capacity for progressive enhancement of user profile quality and, consequently, knowledge tracing prediction accuracy. This stage implements an iterative refinement loop where the Analyst is optimized using feedback from the Predictor's performance. This optimization is guided by Kahneman-Tversky Optimization (KTO) principles~\cite{ethayarajh2024kto}, which leverage binary feedback indicating whether a generated profile contributes to an accurate prediction by the Predictor.

The iterative cycle unfolds as follows:

\begin{enumerate}[leftmargin=*, topsep=0pt, partopsep=0pt, itemsep=0pt, parsep=0pt]
    \item \textbf{Profile Generation:} The current Analyst, parameterized by $\theta_A$ and denoted as a policy $\pi_{\theta_A}$, generates a user profile $\mathbf{p}_t$ from a given student's historical interaction sequence $\mathbf{x}_t$:
    \begin{equation}
        \mathbf{p}_t \sim \pi_{\theta_A}(\cdot \mid \mathbf{x}_t)
        \label{eq:analyst_profile_kto} 
    \end{equation}

    \item \textbf{Prediction and Reward Computation:} The current Predictor (parameterized by $\theta_P$, denoted $f_{\theta_P}$) utilizes $\mathbf{p}_t$, $\mathbf{x}_t$, and potentially the next exercise $e_{t+1}$, to predict student performance $\hat{y}_{t+1}$. This prediction is compared against the ground truth $y_{t+1}$ to yield a binary reward $r_{t+1}$:
    \begin{equation} 
        \hat{y}_{t+1} = f_{\theta_P}(\mathbf{x}_t, \mathbf{p}_t, e_{t+1}) 
    \end{equation}
    \begin{equation}
        r_{t+1} =
        \begin{cases}
            +1, & \text{if } \hat{y}_{t+1} = y_{t+1} \\
            -1, & \text{if } \hat{y}_{t+1} \neq y_{t+1}
        \end{cases}
        \label{eq:kto_reward} 
    \end{equation}

    \item \textbf{Analyst Optimization:} The reward $r_{t+1}$ guides the update of the Analyst's parameters $\theta_A$, encouraging the generation of profiles that lead to accurate Predictor outcomes. The KTO loss function for a batch of instances is:
    \begin{equation}
        \mathcal{L}_{\text{KTO}}(\theta_A) = - \sum_{t} \left[ r_{t+1} \cdot \log \pi_{\theta_A}(\mathbf{p}_t \mid \mathbf{x}_t) \right]
        \label{eq:kto_loss} 
    \end{equation}
    where the sum is over instances $t$ in a training batch. This update resembles a policy gradient step, with $\log \pi_{\theta_A}(\mathbf{p}_t \mid \mathbf{x}_t)$ being the log-probability of generating profile $\mathbf{p}_t$.

    \item \textbf{Predictor Re-training:} After the Analyst is updated (to $\theta_A^{\text{updated}}$) and its profiling capabilities are enhanced, the Predictor can be retrained or further fine-tuned. This uses profiles $\mathbf{p}^{\text{new}}$ generated by the improved Analyst and minimizes the predictive loss (Equation~\ref{eq:predictor_loss}):
    \begin{align}
    \label{eq:predictor_retrain_kto} 
    \theta_P \leftarrow \arg\min_{\theta_P} \sum_{(s,t) \in \mathcal{D}_{\text{train}}}
    \mathcal{L}_{\text{CE}}\big( \notag 
    f_{\theta_P}(\mathcal{H}_{s,t-1},\\\, \mathbf{p}^{\text{new}}_{s,t-1},\, e_t),\, y_{s,t} \big)
    \end{align}
    where $\mathbf{p}^{\text{new}}_{s,t-1} = \text{Analyst}(\mathcal{H}_{s,t-1}; \theta_A^{\text{updated}})$.
\end{enumerate}
This entire cycle is repeated iteratively, fostering mutual improvements in both the Analyst and Predictor components.
\section{Experiments}

To systematically evaluate the efficacy, robustness, and contributions of key components within our proposed collaborative knowledge tracing framework based on large language models, this section details a series of comprehensive experiments. These experiments are designed to thoroughly investigate and address the following core research questions:

\begin{itemize}[leftmargin=*, topsep=0pt, partopsep=0pt, itemsep=0pt, parsep=0pt] 
    \item \textbf{RQ1: Overall Performance}\\
    Can our CIKT surpass traditional knowledge tracing methods and other state-of-the-art large language model baselines?
    \item \textbf{RQ2: Ablation Study} \\
    What are the specific impacts of core design elements affecting overall predictive performance? What are the actual contributions of each component?
    \item \textbf{RQ3: Sensitivity Analysis}\\
    How do two critical parameters, the total number of iteration rounds and the sample size used per iteration, concretely affect the final predictive performance, and what degree of sensitivity does the model exhibit to variations in these parameters?
    \item \textbf{RQ4: Explainability}\\
    Do the user profiles generated by the Analyst demonstrably enhance the explainability of student knowledge states, and do they offer effective guidance for the Predictor's subsequent outcome predictions?
\end{itemize}

\subsection{Experimental Setup}

\subsubsection{Datasets}
Our framework was evaluated on three widely used public educational datasets, providing extensive student interaction records for robust KT model training and validation:

\begin{itemize}[leftmargin=*, topsep=0pt, partopsep=0pt, itemsep=0pt, parsep=0pt]
    \item \textbf{ASSIST09}~\cite{feng2009addressing}: Collected from the ASSISTments online mathematics tutoring system (2009-2010). We utilized the combined version, common in KT research.
    \item \textbf{ASSIST12}~\cite{feng2009addressing}: Also from the ASSISTments platform, this dataset contains student interaction data from 2012-2013.
    \item \textbf{Eedi}~\cite{wang2020instructions}: Sourced from the Eedi mathematics platform as part of the NeurIPS 2020 Education Challenge. We used the \textit{train\_task\_1\_2.csv} file from this challenge, adopting the leaf nodes of its provided math concept tree as the relevant knowledge concepts (KCs) for each question.
\end{itemize}

Prior to model training, the datasets underwent the following preprocessing steps:
\begin{itemize}[leftmargin=*, topsep=0pt, partopsep=0pt, itemsep=0pt, parsep=0pt]
    \item \textbf{Data Cleaning:} Invalid or duplicate records, such as those lacking essential information (e.g., student/question IDs, response correctness), were removed.
    \item \textbf{User Interaction Sequence Construction:} Student records were organized into chronological interaction sequences, each detailing the assessed knowledge concepts (KCs), question difficulty, and response correctness.
    \item \textbf{Question Difficulty Calculation:} Question difficulty was estimated as $1 - \text{pass rate}$, derived from the average correctness for each question within the training set.
    \item \textbf{Sequence Segmentation and Filtering:} Complete student interaction sequences were segmented into 50-interaction subsequences; those with fewer than 5 interactions were removed to ensure effective modeling length.
\end{itemize}
Detailed statistics of the preprocessed datasets are summarized in Table~\ref{tab:dataset_statistics}.

\subsubsection{Backbone}
\label{sec:backbone_llms}
In our framework, both the Analyst and Predictor components leverage large language models as their backbone. We experimented with two primary models: Llama3.1-8B-Instruct and Qwen2.5-7B-Instruct. All experiments reported in this paper were conducted on a single NVIDIA A100 GPU.

\subsubsection{Evaluation}
For model evaluation, each dataset was partitioned into training, validation, and test sets using an 8:1:1 ratio. 
As KT is a binary classification task, we employed ACC and the F1-score as standard evaluation metrics. 
To specifically assess framework scalability and performance on longer interaction sequences, we additionally computed ACC for subsequences where the total number of interactions (including the item to be predicted) exceeds 15, denoted as $\text{ACC}_{len>15}$. The model achieving the best validation set performance was subsequently used for test set evaluation, with prediction results assessed at the final position of each processed sequence. We take the average of five times on the test set as the displayed result. The detailed impact of sequence length characteristics, including the $\text{ACC}_{len>15}$ metric, on model performance is further discussed in the RQ1: Overall Performance section.

\subsubsection{Baselines}
\label{sec:baselines}
To comprehensively evaluate our CIKT framework, we compare its performance against a diverse set of nine baselines, covering both mainstream DLKT methods and general-purpose LLMs. The selected DLKT models include pioneering approaches such as DKT~\cite{piech2015deep} and the memory-augmented DKVMN~\cite{zhang2017dynamic}; Transformer-based architectures like SAKT~\cite{pandey2019self} and AKT~\cite{ghosh2020context}; LPKT~\cite{shen2021learning}, which explicitly models learning and forgetting dynamics; and other established methods IKT~\cite{minn2022interpretable} and DIMKT~\cite{shen2022assessing}. This group represents a spectrum of well-regarded techniques in the KT field. Furthermore, to benchmark against general LLM capabilities when applied directly to the knowledge tracing task, we include GPT-4o~\cite{hurst2024gpt} and Deepseek R1~\cite{guo2025deepseek} as LLM baselines.

\subsection{RQ1: Overall Performance}

To address our first research question (RQ1) concerning the predictive efficacy of our proposed framework, this section compares our CIKT against traditional KT and several LLM baselines. Detailed performance metrics across the ASSIST2009, ASSIST2012, and Eedi are presented in Table~\ref{tab:main}.

Our CIKT framework, particularly the CIKT-Qwen2.5-7B variant, demonstrates highly competitive performance, significantly surpassing the strongest traditional KT baselines in terms of Accuracy (ACC) and F1-score across all three datasets. Notably, this performance advantage of CIKT is often more pronounced on longer interaction sequences. Such superior performance on extended histories underscores CIKT's enhanced capability to effectively model long-range dependencies and leverage comprehensive contextual information through its dynamic student profiling mechanism. Furthermore, CIKT showcases a clear advantage over general-purpose LLM baselines like GPT-4o and Deepseek-R1 when applied directly to the KT task. Their considerably lower performance highlights the necessity of CIKT's specialized, structured, iterative, and collaborative architecture, which features explicit student profile generation and targeted optimization, as opposed to direct, unspecialized LLM application.

These findings effectively address RQ1, demonstrating that our framework achieves superior performance in knowledge tracing compared to both traditional methods and direct LLM applications.

\begin{figure*}[htbp]
  \centering
  \includegraphics[width=1.0\textwidth]{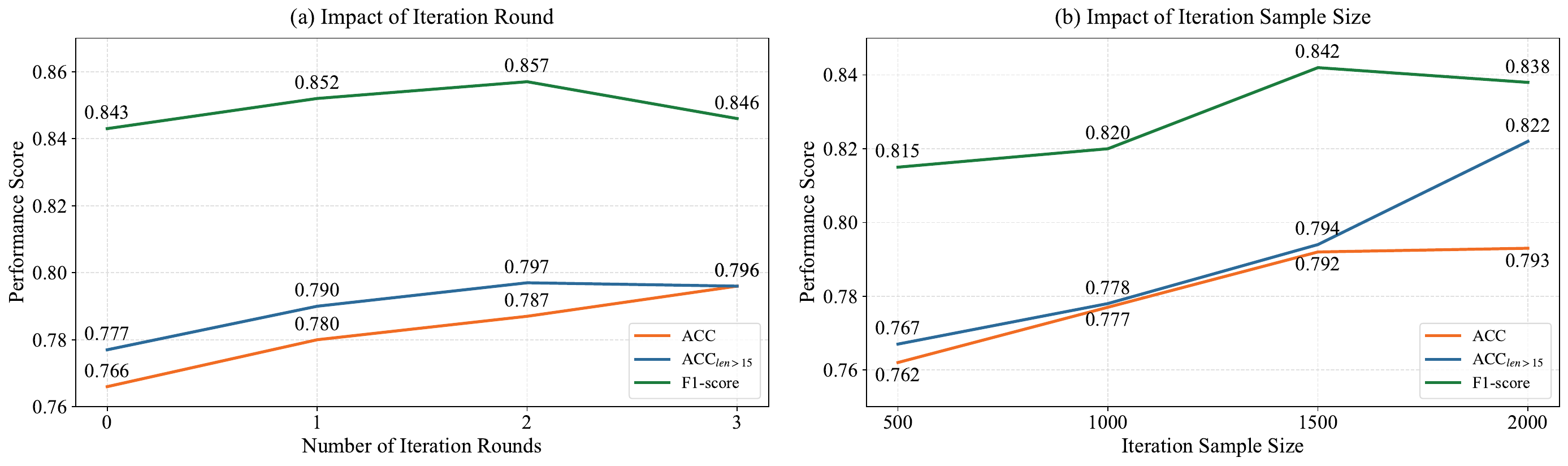}
  \caption{Effect of the number of interaction rounds and sample size per iteration on the performance of CIKT.}
  \label{fig:parameter}
\end{figure*}

\subsection{RQ2: Ablation Study}

To investigate the individual contributions of key components within our CIKT framework—namely, the iterative optimization, the cooperative Analyst-Predictor structure with profile generation, and the utilization of profiles at inference—we conducted a series of ablation studies, thereby addressing RQ2. These experiments were performed using both Llama3.1-8B-Instruct and Qwen2.5-7B-Instruct as backbone models. The specific configurations and detailed results of these studies are presented in Table~\ref{tab:ablation}, with the caption of the table defining each ablated variant.

The results in Table~\ref{tab:ablation} consistently demonstrate the critical importance of each evaluated component across both backbone models and all datasets. Comparing the full CIKT framework to \textbf{CIKT w/o Iteration}, a clear performance drop is observed, highlighting the positive impact of the iterative process on refining profile quality and enhancing Analyst-Predictor synergy. When both iteration and the cooperative profiling structure are removed (\textbf{CIKT w/o Iteration \& Cooperation}), representing a direct LLM fine-tuning approach, performance generally degrades further. This underscores the benefits of CIKT's explicit two-component architecture and profile-based modeling, even without iterative refinement. Most notably, the \textbf{CIKT w/o Profile (Inference)} configuration, where profiles are withheld from the Predictor during inference after full CIKT training, results in the most significant performance deterioration. This unequivocally confirms the crucial role of the dynamically generated student profiles in enabling accurate predictions, as their absence severely hampers the Predictor's capability. The results address RQ2.

\subsection{RQ3: Sensitivity Analysis}

To address RQ3, we analyzed model sensitivity to the total number of iteration rounds and the iteration sample size, denoted as $k$. These results are visually summarized in Figure~\ref{fig:parameter}.

First, our examination of iteration rounds, varied from 0 to 3 for the CIKT-Qwen2.5-7B model on the ASSIST2012 dataset and depicted in Figure~\ref{fig:parameter}(a), revealed that performance generally stabilized or reached a strong point around three iterations. This underscores the efficacy of progressive refinement in enhancing the model's predictive capabilities.

Next, regarding sensitivity to the iteration sample size $k$, which was varied from 500 to 2000 using CIKT-Qwen2.5-7B on the ASSIST2009 dataset as shown in Figure~\ref{fig:parameter}(b), results indicated that while larger $k$ values generally improved overall metrics, these gains diminished at the higher end of the tested range. Notably, ACC on longer sequences, specifically the $\text{ACC}_{len>15}$ metric for sequences with more than 15 interactions, exhibited more sustained improvement with increasing $k$. This suggests a particular advantage for modeling long-term dependencies. Considering the trade-off between overall efficacy and computational overhead, an iteration sample size of $k=1000$ was adopted for most other experiments.

\subsection{RQ4: Explainability and Profile Utility}
\label{sec:rq4_explainability_utility}

RQ4 assesses the explainability offered by our CIKT framework and the predictive utility of the user profiles generated by the Analyst. These dynamic textual profiles are designed to provide human-understandable summaries of student learning patterns, concept mastery, and areas needing attention. Due to space constraints, a detailed qualitative case study is presented in Section~\ref{sec:Case Study}. This case study illustrates how the profiles are refined through the iterative KTO process to yield more nuanced pedagogical insights and how these improved profiles contribute to the Predictor's enhanced, more interpretable forecasting, thereby addressing both aspects of RQ4.

\section{Conclusion}
\label{sec:conclusion}

We proposed CIKT to iteratively optimize student profiling and performance prediction for accurate and explainable knowledge tracing. Its synergistic architecture enables continuous mutual enhancement, yielding significantly improved predictive accuracy and explainable student profiles on multiple educational datasets.

\section*{Limitations}

Because our framework primarily generates binarized judgments for student responses, we focused on metrics such as ACC and F1-score, and consequently did not employ ranking-sensitive evaluation metrics like AUC.
Moreover, due to the context window constraints of our backbone models, we did not incorporate the textual content of question stems, which limited the potential for more fine-grained, content-aware modeling of student knowledge.

\bibliography{custom}

\appendix

\section{Notation Table}
\label{sec:Notation Table}
We list and explain the notations in our methodology in Table~\ref{tab:notation_cikt}.

\begin{table*}[htbp]
\centering
\renewcommand{\arraystretch}{1.2} 
\begin{tabular}{ll}
\hline
\textbf{Notation} & \textbf{Description} \\
\hline
\multicolumn{2}{l}{\textbf{General Notations}} \\
$s$ & Index for a student \\
$e_i$ & The $i$-th exercise or question encountered by a student \\
$r_i$ & Binary correctness ($0$ or $1$) of the student's response to $e_i$ \\
$\mathcal{S}_s$ & Historical interaction sequence for student $s$, e.g., $\mathcal{S}_s = \{(e_1, r_1), \dots, (e_N, r_N)\}$ \\
$\mathbf{p}_s$ & User profile (typically textual) for student $s$, generated by the Analyst \\
$\theta_A$ & Parameters of the Analyst model \\
$\theta_P$ & Parameters of the Predictor model \\
$\mathcal{D}_{\text{train}}$ & Set of training instances (e.g., students or student-interaction sequences) \\
$\mathcal{L}_{\text{CE}}(\cdot, \cdot)$ & Standard cross-entropy loss function \\
\hline
\multicolumn{2}{l}{\textbf{Stage 1: Distillation}} \\
$\text{LLM}_{\text{teacher}}$ & Large-parameter teacher model (e.g., GPT-4o) for initial profile annotation \\
$\mathbf{p}_{s, \text{teacher}}$ & Initial textual profile for student $s$ generated by $\text{LLM}_{\text{teacher}}$ (Eq.~\ref{eq:raw_teacher_profile}) \\
$\mathbf{p}^*_{s, \text{teacher}}$ & Curated, high-quality teacher profile for student $s$ used for training \\
$\mathbf{p}_{s, \text{analyst}}$ & Profile for student $s$ generated by the Analyst during distillation (Eq.~\ref{eq:analyst_profile_distill}) \\
$\mathcal{L}_{\text{Distill}}(\theta_A)$ & Distillation loss for training the Analyst (Eq.~\ref{eq:distill_loss_ce}) \\
\hline
\multicolumn{2}{l}{\textbf{Stage 2: Profiling}} \\
$\{\mathbf{p}_s\}$ & Set of all generated user profiles \\
$\{\mathcal{S}_s\}$ & Set of all student interaction sequences \\
\hline
\multicolumn{2}{l}{\textbf{Stage 3: Reasoning}} \\
$\mathcal{H}_{s,t-1}$ & Historical interaction sequence for student $s$ up to interaction $t-1$ \\
$\mathbf{p}_{s,t-1}$ & User profile for student $s$ based on history $\mathcal{H}_{s,t-1}$ \\
$e_t$ & The $t$-th (current or next) exercise for student $s$ \\
$y_{s,t}$ & Actual binary correctness of student $s$'s response to $e_t$ \\
$\hat{y}_{s,t}$ & Predicted probability of student $s$ correctly answering $e_t$ (Eq.~\ref{eq:predictor_output}) \\
$\mathcal{L}_{\text{Predict}}(\theta_P)$ & Predictive loss for training the Predictor (Eq.~\ref{eq:predictor_loss}) \\
\hline
\multicolumn{2}{l}{\textbf{Stage 4: Iteration}} \\
$\mathbf{x}_t$ & A student's historical interaction sequence  \\
$\pi_{\theta_A}$ & The Analyst viewed as a policy parameterized by $\theta_A$ \\
$\mathbf{p}_t$ & User profile generated by Analyst $\pi_{\theta_A}$ from $\mathbf{x}_t$ (Eq.~\ref{eq:analyst_profile_kto}) \\
$f_{\theta_P}$ & The Predictor model parameterized by $\theta_P$ \\
$e_{t+1}$ & The subsequent exercise for which a prediction is made based on $\mathbf{x}_t, \mathbf{p}_t$ \\
$y_{t+1}$ & Ground truth outcome for the prediction $\hat{y}_{t+1}$ \\
$\hat{y}_{t+1}$ & Prediction by $f_{\theta_P}$ for $e_{t+1}$ \\
$r_{t+1}$ & Binary reward signal ($+1$ or $-1$) based on prediction accuracy (Eq.~\ref{eq:kto_reward}) \\
$\mathcal{L}_{\text{KTO}}(\theta_A)$ & KTO loss function for optimizing the Analyst (Eq.~\ref{eq:kto_loss}) \\
$\log \pi_{\theta_A}(\mathbf{p}_t \mid \mathbf{x}_t)$ & Log-probability of the Analyst generating profile $\mathbf{p}_t$ given sequence $\mathbf{x}_t$ \\
$\theta_A^{\text{updated}}$ & Updated parameters of the Analyst after a KTO step \\
$\mathbf{p}^{\text{new}}_{s,t-1}$ & Profile generated by the updated Analyst using history $\mathcal{H}_{s,t-1}$ \\
\hline
\end{tabular}
\caption{Notation Table for the CIKT Framework.}
\label{tab:notation_cikt}
\end{table*}

\section{Hyper-parameter Setting}

We provide the training and inference hyper-parameter settings in Table~\ref{tab:Hyper-parameter-training} and Table~\ref{tab:Hyper-parameter-inference}.

\begin{table}[htbp]
\centering
\small
\begin{tabular}{lcc}
\hline
\textbf{Parameter(training)} & Analyst & Predictor\\
\hline
\textbf{\# lora\_rank} & 8 & 8 \\
\textbf{\# learning\_rate} & 5.0e-6 & 1.0e-4 \\
\textbf{\# train\_epochs} & 10 & 10 \\
\textbf{\# warmup\_ratio} & 0.1 & 0.1 \\
\hline
\end{tabular}
\caption{Training hyper-parameter setting of CIKT.}
\label{tab:Hyper-parameter-training}
\end{table}

\begin{table}[htbp]
\centering
\small
\begin{tabular}{lcc}
\hline
\textbf{Parameter(inference)} & Analyst & Predictor\\
\hline
\textbf{\# temperature} & 0.95 & 0 \\
\textbf{\# top\_p} & 0.7 & 0.7 \\
\textbf{\# top\_k} & 50 & 50 \\
\hline
\end{tabular}
\caption{Inference hyper-parameter setting of CIKT.}
\label{tab:Hyper-parameter-inference}
\end{table}

\section{Case Study}
\label{sec:Case Study}

To qualitatively illustrate the efficacy of our CIKT framework's iterative optimization process, particularly its impact on the quality of student profiles generated by the Analyst and the subsequent prediction accuracy of the Predictor, we present a detailed case study. The selected case, detailed in Table~\ref{tab:casestudy}, involves a student interaction sequence where the Predictor's outcome for the "Next Question" was incorrect based on the initial profile generated by the Analyst before iteration, but became correct after the Analyst was refined through iterations. The ground truth for the "Next Question" -- (['Conversion of Fraction Decimals Percents'], 0.16) -- was "False".

\textbf{Profile Before Iteration and Initial Prediction.}
The initial user profile generated by the Analyst before iterative refinement is presented in the left column of the bottom table in Table~\ref{tab:casestudy}. This profile, while attempting to summarize performance across concepts like "Making a Table from an Equation," "Equivalent Fractions," and "Conversion of Fraction Decimals Percents," tends to exhibit characteristics of a more direct translation or surface-level summary of the interaction sequence. For instance, it meticulously lists the number of attempts and correctness for each topic (e.g., "The student has encountered three questions related to this topic, all answered incorrectly" for "Making a Table from an Equation"). While it provides some overview, it contains redundancies in its factual recounting and offers a somewhat limited depth of analytical insight beyond stating observed patterns.

Crucially, regarding the "Next Question" on "Conversion of Fraction Decimals Percents" (difficulty 0.16), the profile (as highlighted in Red notes: "\textit{- This question is slightly easier than the previous one, which may provide an opportunity for the student to consolidate their understanding.}" This particular phrasing, emphasizing the "easier" nature and "opportunity to consolidate," might lead the Predictor to infer a higher likelihood of a correct answer. In this instance, the Predictor, relying on this pre-iteration profile, made an incorrect prediction (implicitly predicting "True", while the real response was "False").

\textbf{Profile After Iteration and Corrected Prediction.}
The right column of the bottom table in Table~\ref{tab:casestudy} showcases the student profile generated by the Analyst after several KTO iterations. This refined profile demonstrates a notable improvement in several aspects. It is more focused in its analysis, moving beyond simple sequence translation to offer more structured insights and actionable recommendations. For example, it categorizes observations into "Overall Performance and Patterns," "Difficulty and Learning Progression," "Projected Next Question," and detailed "Recommendations." This structure itself provides a clearer, more pedagogically useful summary. The suggestions provided, such as "Focus on reinforcing understanding of "Table" as a standalone topic..." and "Work on integrating concepts...", are more specific and reliable, offering genuine guidance applicable in real educational scenarios.

The shift in the analysis of the "Next Question" is particularly significant. The refined profile highlighted in Green states: "\textit{- The upcoming question on "Conversion of Fraction Decimals Percents" with a difficulty of 0.16 is consistent with the difficulty level at which the student has previously struggled with this knowledge point, presenting a challenge.}" This revised perspective, informed by the iterative feedback loop, correctly identifies the question as a challenge despite its slightly lower difficulty, considering the student's prior struggles with the same concept at a similar difficulty (0.17, as per the profile's detailed breakdown). This more nuanced and context-aware assessment likely guided the Predictor to correctly forecast the outcome as "False" for the "Next Question", aligning with the ground truth.

\begin{table*}[htbp]
    \centering
    \renewcommand{\arraystretch}{1.3}
    \small
    \begin{tabular}{p{0.96\linewidth}}  
        \hline
        \textbf{Question and Input} \\
        \hline
        \textbf{Question:} \newline
        The student has completed some questions in a listing order from earliest to latest, with each ternary in the subsequent sequence representing a piece of information containing the knowledge points contained in the question, the difficulty of the question (expressed as the complement of the pass rate on the full set), and the response to the question, whether true or false. The sequence culminates with the presentation of a binary comprising the knowledge points and the difficulty of the subsequent question. Please provide a high-level summary of the student's state of knowledge acquisition in detail.
        \newline
        \textbf{Input:} \newline
        The student's historical response sequence: \newline
        (['Making a Table from an Equation'], 0.47, False), (['Making a Table from an Equation'], 0.64, False), (['Making a Table from an Equation'], 0.61, False), (['Equivalent Fractions'], 0.13, True), (['Equivalent Fractions'], 0.26, True), (['Equivalent Fractions'], 0.19, True), (['Equivalent Fractions'], 0.37, False), (['Equivalent Fractions'], 0.32, False), (['Equivalent Fractions'], 0.39, False), (['Conversion of Fraction Decimals Percents'], 0.17, True), (['Conversion of Fraction Decimals Percents'], 0.17, False)
        \newline
        \textbf{Next Question:} (['Conversion of Fraction Decimals Percents'], 0.16) \newline
        \textbf{Real response(No input):} False \\
    \end{tabular}


    \begin{tabular}{p{0.48\linewidth} | p{0.48\linewidth}}
        \hline
        \textbf{Profile Before Iteration} & \textbf{Profile After Iteration} \\
        \hline
        1. \textbf{Making a Table from an Equation}: \newline 
        \hspace*{1em}- The student has encountered three questions related to this topic, all answered incorrectly. \newline
        \hspace*{1em}- The difficulty levels of these questions ranged from 0.47 to 0.64, indicating a moderate level of challenge. \newline
        \hspace*{1em}- This suggests that the student may not yet have a solid understanding of this concept or is struggling with the application of this knowledge. 
        \newline
        2. \textbf{Equivalent Fractions}: \newline
        \hspace*{1em}- The student has answered five questions related to this topic, with a mix of correct and incorrect responses. \newline
        \hspace*{1em}- The difficulty levels ranged from 0.13 to 0.39. \newline
        \hspace*{1em}- The student demonstrated a good grasp of easier questions (e.g., those with a lower difficulty, such as 0.13 and 0.26), but struggled with more challenging ones (e.g., 0.37, 0.32, and 0.39). \newline
        \hspace*{1em}- This indicates that while the student has some understanding of the topic, there may be gaps in knowledge or difficulty with more complex problems. 
        \newline
        3. \textbf{Conversion of Fraction Decimals Percents}: \newline 
        \hspace*{1em}- The student has attempted two questions related to this topic. \newline
        \hspace*{1em}- The first question was answered correctly, and the second was answered incorrectly. \newline
        \hspace*{1em}- Both questions had a difficulty level of 0.17. \newline
        \hspace*{1em}- The student shows potential understanding of this topic but may need further practice or reinforcement to solidify their knowledge. \newline
        \textbf{Next Question}: \newline 
        \hspace*{1em}- The next question is related to "Conversion of Fraction Decimals Percents" with a difficulty level of 0.16. \newline
        \sethlcolor{myRed} 
        \hspace*{1em}\hl{- This question is slightly easier than the previous one, which may provide an opportunity for the student to consolidate their understanding.} \newline
        \textbf{Overall Summary}: \newline 
        \hspace*{1em}- The student demonstrates varying levels of understanding across different topics. \newline
        \hspace*{1em}- They show strong potential in "Equivalent Fractions" for easier questions but may need more practice with more challenging problems. \newline
        \hspace*{1em}- They seem to have a good understanding of "Conversion of Fraction Decimals Percents" but need reinforcement. \newline
        \hspace*{1em}- There is a clear area for improvement in "Making a Table from an Equation," as the student has consistently answered questions incorrectly, indicating a need for focused practice and possibly revisiting foundational concepts. 
         &
        1. \textbf{Making a Table from an Equation:} \newline
        \hspace*{1em}- The student initially struggled with questions related to "Making a Table from an Equation," as evidenced by incorrect responses to questions with moderate difficulty levels (0.47, 0.64, and 0.61). This suggests a gap in understanding or application of this concept. \newline
        2. \textbf{Equivalent Fractions:} \newline
        \hspace*{1em}- The student demonstrated initial proficiency with "Equivalent Fractions," correctly answering questions with low to moderate difficulty levels (0.13 and 0.26). However, there was a decline in performance as the difficulty increased, with incorrect responses to questions at 0.37, 0.32, and 0.39. This indicates that while the student has a foundational understanding, they may struggle with more complex applications or problem variations of this concept. 
        \newline
        3. \textbf{Conversion of Fraction Decimals Percents:} \newline 
        \hspace*{1em}- The student showed mixed results with "Conversion of Fraction Decimals Percents." They correctly answered a question with a low difficulty level (0.17), indicating a basic understanding, but then responded incorrectly to another question of the same difficulty level (0.17). This inconsistency suggests that the student may need further practice or clarification to solidify their understanding of this concept. \newline
        4. \textbf{Overall Knowledge Acquisition:} \newline
        \hspace*{1em}- The student's knowledge acquisition shows a pattern of initial difficulty with a concept, followed by some improvement or partial understanding, but often ending with inconsistencies or gaps in understanding, especially as the difficulty increases. \newline
        \sethlcolor{myGreen} 
        \hspace*{1em}\hl{- The upcoming question on "Conversion of Fraction Decimals Percents" with a difficulty of 0.16 is consistent with the difficulty level at which the student has previously struggled with this knowledge point, presenting a challenge.}
        \newline
        5. \textbf{Recommendations:} \newline
        \hspace*{1em}- The student should focus on reinforcing their understanding of "Making a Table from an Equation" and "Equivalent Fractions" through targeted practice and review of fundamental concepts. \newline
        \hspace*{1em}- For "Conversion of Fraction Decimals Percents," the student should review the basic principles and practice a variety of problems to build confidence and consistency in their responses. \newline
        \hspace*{1em}- Providing feedback on incorrect responses and ensuring the student understands the reasoning behind the correct answers will be crucial in improving their overall performance.\newline
        - Gradually increase difficulty level in practiced areas.
        \\
        \hline
    \end{tabular}
    \caption{Example of profiles before and after iteration.}
    \label{tab:casestudy}
\end{table*}

\end{document}